\documentclass[11pt]{article}

\usepackage[final]{acl}
\usepackage{times}
\usepackage{latexsym}
\usepackage[T1]{fontenc}
\usepackage[utf8]{inputenc}
\usepackage{microtype}
\usepackage{inconsolata}
\usepackage{graphicx}
\usepackage{xurl}

\title{LLM-to-Speech: A Synthetic Data Pipeline for Training Dialectal Text-to-Speech Models}

\author{Ahmed Khaled Khamis \\
  Georgia Institute of Technology \\
  \texttt{akhamis6@gatech.edu}
  \And
  Hesham Ali \\
  Nile University \\
  \texttt{he.ali@nu.edu.eg}
}

\begin{document}
\maketitle
\begin{abstract}
Despite the advances in neural text to speech (TTS), many Arabic dialectal varieties remain marginally addressed, with most resources concentrated on Modern Spoken Arabic (MSA) and Gulf dialects, leaving Egyptian Arabic---the most widely understood Arabic dialect--- severely under-resourced. We address this gap by introducing NileTTS: 38 hours of transcribed speech from two speakers across diverse domains including medical, sales, and general conversations. We construct this dataset using a novel synthetic pipeline: large language models (LLM) generate Egyptian Arabic content, which is then converted to natural speech using audio synthesis tools, followed by automatic transcription and speaker diarization with manual quality verification. We fine-tune XTTS v2, a state-of-the-art multilingual TTS model, on our dataset and evaluate against the baseline model trained on other Arabic dialects. Our contributions include: (1) the first publicly available Egyptian Arabic TTS dataset, (2) a reproducible synthetic data generation pipeline for dialectal TTS, and (3) an open-source fine-tuned model. All resources are released to advance Egyptian Arabic speech synthesis research.
\end{abstract}

\section{Introduction}
Neural text-to-speech (TTS) has made remarkable progress in recent years, with models like Tacotron \cite{tacotron}, FastSpeech \cite{fastspeech}, and VITS \cite{vits} achieving near-human naturalness for high-resource languages. More recently, multilingual TTS systems such as XTTS \cite{xtts} and VALL-E \cite{valle} have demonstrated impressive zero-shot voice cloning capabilities across different languages. However, this progress has not been evenly distributed, as low-resource languages and dialectal varieties remain significantly under-served.

Arabic presents a particularly challenging case for TTS research. While Modern Standard Arabic (MSA) has received considerable attention, the spoken reality of the Arab world is far more diverse. Arabic has many regional dialects that differ substantially in phonology, vocabulary, and syntax, often to the point of mutual unintelligibility \cite{arabic-dialects}. Among these, Egyptian Arabic holds a unique position: spoken natively by over 100 million people and widely understood across the Arab world due to Egypt's dominant media presence, it is arguably the most accessible Arabic variety.

Despite its prominence, Egyptian Arabic remains under-resourced for speech synthesis. While prior work has explored Egyptian Arabic TTS \cite{masry}, \cite{sawtarabi}, existing resources are limited in scale, domain coverage, or public availability. Current Arabic TTS systems mainly target MSA or Gulf dialects, leaving Egyptian Arabic speakers without state of the art tools. As a result, Egyptian Arabic speakers lack access to quality TTS in applications like voice assistants and audiobooks.

In this work, we address this resource gap by introducing NileTTS\footnote{Code: \url{https://github.com/KickItLikeShika/NileTTS}} , a large-scale Egyptian Arabic TTS dataset along with a fine-tuned speech synthesis model. Our dataset comprises 38 hours of transcribed Egyptian Arabic speech from two speakers across three domains: medical, sales and customer service, and general conversation.

A key contribution of our work is the novel synthetic data generation pipeline used to construct the dataset. Rather than relying on costly manual recording, we leverage recent advances in generative AI: large language models (LLMs) generate Egyptian Arabic content across diverse topics, which is then converted to natural-sounding speech using neural audio synthesis tools that support Egyptian Arabic. The resulting audio is automatically transcribed using Whisper \cite{whisper} and segmented into utterances, with speaker identities assigned via \textit{ECAPA-TDNN-based} speaker diarization \cite{ecapa}. Manual quality verification ensures transcription accuracy and speaker consistency. This pipeline offers a reproducible and scalable approach for creating TTS datasets for other low-resource dialects.

To demonstrate the utility of our dataset, we fine-tune XTTS v2 \cite{xtts}, a state-of-the-art multilingual TTS model with zero-shot voice cloning capabilities. We evaluate the fine-tuned model against the baseline XTTS v2, which was trained on Arabic data from other dialectal varieties. Our experiments show substantial improvements in intelligibility and speaker similarity.
Our contributions are as follows:
\begin{itemize}
\item We release \textbf{NileTTS}\footnote{Dataset: \url{https://huggingface.co/datasets/KickItLikeShika/NileTTS-dataset}}, a large-scale Egyptian Arabic TTS dataset comprising 38 hours of transcribed speech across multiple domains.
\item We present a \textbf{reproducible synthetic data generation pipeline} combining LLM-based content generation, neural audio synthesis, automatic transcription, and speaker diarization.
\item We provide an \textbf{open-source fine-tuned XTTS model}\footnote{Model: \url{https://huggingface.co/KickItLikeShika/NileTTS-XTTS}} for Egyptian Arabic, serving as a baseline for future research.
\end{itemize}
We publicly release all resources to facilitate further research in Egyptian Arabic speech synthesis.

\section{Related Work}
\subsection{Arabic Text-to-Speech}
Arabic TTS research has primarily focused on Modern Standard Arabic (MSA), with systems leveraging both traditional concatenative methods and neural approaches \cite{sawtarabi}. For dialectal Arabic, resources remain scarce. Notable exceptions include work on Gulf Arabic dialects, which benefit from commercial interest in the Gulf region.

For Egyptian Arabic specifically, two prior efforts are most relevant. \citet{masry} introduced EGYARA-23, a 20.5-hour dataset featuring a single male speaker narrating news and general conversations, comprising 32,716 segments. While substantial in size, the dataset is limited to one speaker and two domains. More recently, \citet{sawtarabi} presented SawtArabi, a multi-dialect Arabic speech corpus that includes approximately one hour of Egyptian Arabic among several other varieties. While valuable for cross-dialectal research, the Egyptian Arabic portion is limited in scale for dedicated TTS training.

Our work complements these efforts by providing a larger, more diverse resource: 38 hours of Egyptian Arabic speech from two speakers (male and female) across three distinct domains. Additionally, we introduce a synthetic data generation pipeline that offers a reproducible approach for future dataset expansion.

\subsection{Synthetic Data for Speech}
Synthetic data generation has emerged as a promising approach for low-resource speech tasks. Prior work has explored using TTS systems to generate training data for automatic speech recognition \cite{synth-asr}, and text augmentation via LLMs has shown success in NLP tasks \cite{llm-aug}. Our work extends this paradigm to TTS dataset construction, using LLMs for content generation and neural audio synthesis for speech production—creating a fully synthetic pipeline that requires no manual recording.

\subsection{Multilingual TTS and XTTS}
Recent advances in multilingual TTS have enabled models to synthesize speech across many languages from a single model. XTTS v2 \cite{xtts}, built on a GPT-style architecture with voice cloning capabilities, supports over 16 languages including Arabic. However, its Arabic training data primarily covers MSA and Gulf dialects. We finetune XTTS v2 on our Egyptian Arabic dataset to adapt it to this under-served variety.

\section{Dataset Construction}
\begin{figure*}[t]
\centering
\includegraphics[width=\textwidth]{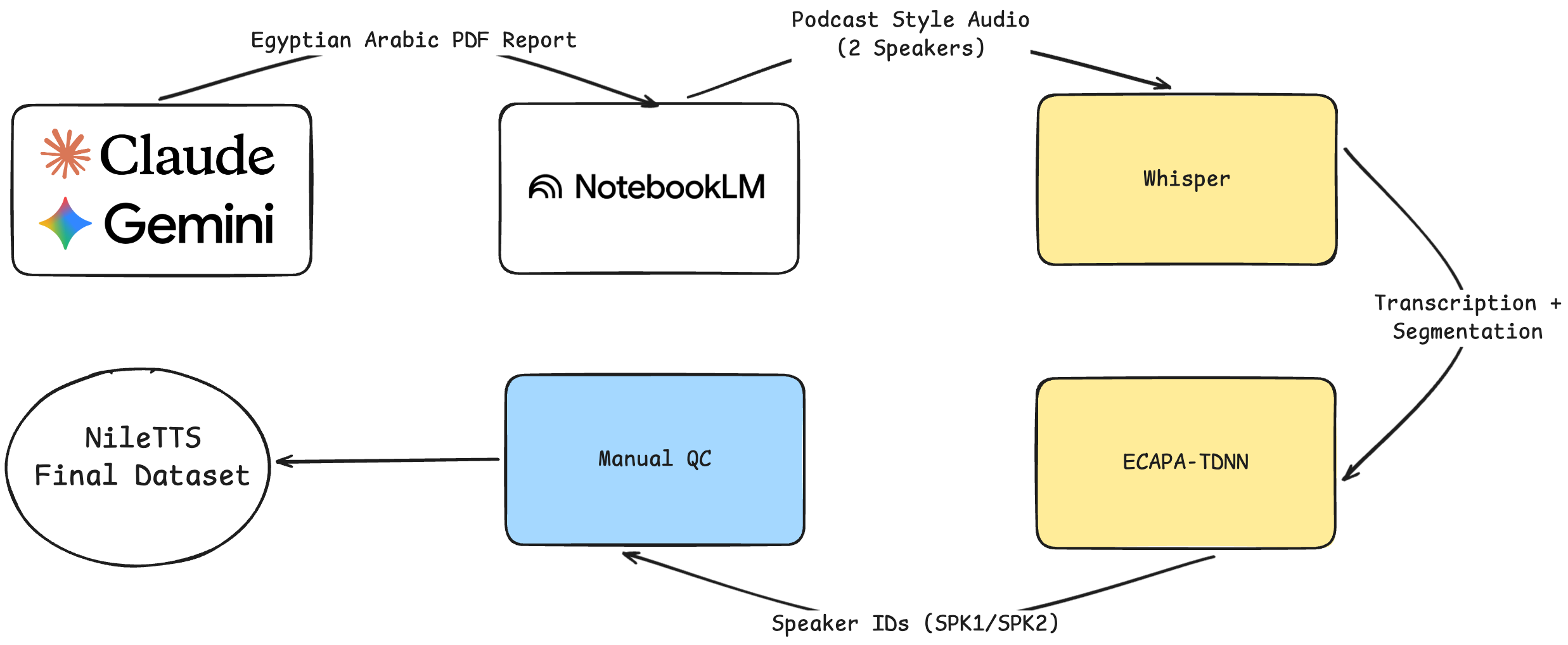}
\caption{Overview of the NileTTS data generation pipeline. Egyptian Arabic content is generated by LLMs, converted to speech via neural audio synthesis, transcribed and segmented using Whisper, and annotated with speaker identities using ECAPA-TDNN embeddings. Manual quality control ensures accuracy before final dataset compilation.}
\label{fig:pipeline}
\end{figure*}

This section describes the construction of the NileTTS dataset. We present a synthetic data generation pipeline that leverages large language models for content creation, neural audio synthesis for speech generation, and automatic tools for transcription and speaker identification. Figure \ref{fig:pipeline} illustrates the complete pipeline.

\subsection{Content Generation}
The first stage of our pipeline involves generating Egyptian Arabic textual content using large language models. We employ variants \textbf{\textit{Gemini}} and \textbf{\textit{Claude}} to generate PDF-style reports on diverse topics in authentic Egyptian Arabic dialect.
We target three domains to ensure topical diversity:
\begin{itemize}
\item \textbf{Medical:} Health topics, symptoms, treatments, and medical advice
\item \textbf{Sales and Customer Service:} Product discussions, negotiation scenarios, customer interactions
\item \textbf{General Conversations:} Everyday topics, social commentary, cultural discussions
\end{itemize}
For each generation, we prompt the LLM to write an a report entirely in Egyptian Arabic dialect, explicitly avoiding Modern Standard Arabic. The prompts specify the domain and request natural conversational language that reflects how Egyptians actually speak. This approach yields content that is both topically diverse and linguistically authentic to Egyptian Arabic.

The prompts used for content generation were intentionally \textbf{simple}. We did not use complex prompting strategies or explicit normalization rules; instead, we directly instructed the model to generate reports in Egyptian Arabic while avoiding Modern Standard Arabic. Dialect authenticity was assessed \textbf{qualitatively} through manual inspection. In practice, both models consistently produced fluent Egyptian Arabic content.

\subsection{Audio Synthesis}
The generated textual reports are then converted to speech using \textbf{\textit{NotebookLM's}} audio generation feature. NotebookLM produces podcast-style audio discussions where two virtual hosts engage in an in-depth natural conversation about the input report. Crucially for our purposes, NotebookLM supports high-quality Egyptian Arabic synthesis with authentic dialect pronunciation.
The audio generation produces conversations featuring two distinct speakers: one male and one female voice. Both speakers maintain consistent voice characteristics across all generated audio, which is essential for TTS training data. Each generated audio file is approximately 10-15 minutes in length, covering the content of one PDF report in a conversational format.
We selected NotebookLM for several reasons: (1) it produces natural, conversational Egyptian Arabic speech rather than formal MSA; (2) the two-speaker format provides speaker diversity within a consistent voice identity; and (3) the audio quality is suitable for TTS training without significant artifacts.

\subsection{Transcription and Segmentation}
The generated audio files are processed using OpenAI's Whisper Large model \cite{whisper} for automatic transcription. Whisper provides accurate Arabic transcription with word-level timestamps, which we use for segmentation.
We segment the continuous audio into utterance-level chunks. This constraint ensures manageable sequence lengths. Segmentation is performed at natural speech boundaries (pauses between utterances) using the timestamp information from Whisper. Segments shorter than 1 second or containing only silence are discarded.
Each segment is saved as an individual WAV file along with its corresponding transcription. This produces the paired (audio, text) format required for TTS training.

\subsection{Speaker Diarization}
Since the source audio contains two speakers in conversation, we must identify which speaker produced each segment. We employ a speaker diarization approach based on speaker embeddings.
We use the ECAPA-TDNN model \cite{ecapa} from SpeechBrain \cite{speechbrain} to extract speaker embeddings. ECAPA-TDNN produces a 192-dimensional embedding vector for each audio segment that captures the speaker's voice characteristics independent of linguistic content.
Our diarization process works as follows: 1) \textbf{Embedding Extraction:} We extract ECAPA-TDNN embeddings for all segments across multiple audio files, 2) \textbf{Centroid Computation:} Using K-Means clustering with $k=2$, we identify two cluster centroids representing the two speakers' average voice characteristics, and 3) \textbf{Speaker Assignment:} For each segment, we compute the Cosine Similarity between its embedding and both centroids. The segment is assigned to the speaker whose centroid is closest.

This approach reliably separates the two speakers, as their voice characteristics (male vs. female) are sufficiently distinct in the embedding space. The computed centroids are saved and reused for processing new audio files, ensuring consistent speaker labels across the entire dataset.

\subsection{Quality Control}
While our pipeline is largely automated, we incorporate manual quality control to ensure dataset quality. Human annotators reviewed a the whole category for Sales and Customer Service, along with a sample of the other 2 sections to verify the following: 1) \textbf{Transcription Accuracy:} Checking that the Whisper transcription correctly captures the spoken content, particularly for Egyptian Arabic vocabulary and expressions that may differ from MSA, 2) \textbf{Speaker Consistency:} Verifying that the automated speaker labels correctly identify the speaker in each segment, and 3) \textbf{Audio Quality:} Ensuring segments are free from artifacts, truncation, or overlapping speech.

Segments with significant errors are corrected or removed. This quality control step is essential for maintaining dataset integrity, particularly for dialectal content where automatic tools may have higher error rates than for standard language varieties.

\subsection{Dataset Statistics}
\begin{table}[t]
\centering
\begin{tabular}{lrr}
\hline
\textbf{Statistic} & \textbf{Utterances} & \textbf{Hours} \\
\hline
Total & 9,521 & 38.1 \\
Training Set & 8,571 & -- \\
Evaluation Set & 950 & -- \\
\hline
Sales \& Customer Service & 4,975 & 21.0 \\
General Conversations & 2,979 & 11.2 \\
Medical & 1,567 & 5.9 \\
\hline
SPEAKER\_01 (Male) & 4,865 & -- \\
SPEAKER\_02 (Female) & 4,656 & -- \\
\hline
Average Utterance Length & \multicolumn{2}{c}{14.4 seconds} \\
\hline
\end{tabular}
\caption{NileTTS dataset statistics.}
\label{tab:dataset-stats}
\end{table}
Table \ref{tab:dataset-stats} summarizes the NileTTS dataset. The final dataset comprises 38.1 hours of transcribed Egyptian Arabic speech, totaling 9,521 utterances. We split the data into training (90\%) and evaluation (10\%) sets, ensuring both speakers appear in both splits while keeping specific utterances exclusive to one split. We ensure that there is \textbf{no report-level overlap} between the training and evaluation sets, and that the evaluation set contains \textbf{unseen topics and prompts} not used during training.
The dataset covers three domains: Sales and Customer Service is the largest (4,975 utterances, 21.0 hours), followed by General Conversations (2,979 utterances, 11.2 hours) and Medical (1,567 utterances, 5.9 hours).
Speaker representation is well-balanced, with SPEAKER\_01 (male) contributing 4,865 utterances and SPEAKER\_02 (female) contributing 4,656 utterances. The conversational format naturally produces roughly equal speaking time between both voices. The average utterance length of 14.4 seconds provides sufficient context for TTS training while remaining within typical sequence length constraints.
The dataset is formatted following the XTTS v2 training data specification: each utterance is stored as a WAV file, paired with its transcription and speaker identifier in a metadata CSV file. This ensures direct compatibility with the XTTS fine-tuning pipeline and facilitates reproducibility.

\section{Model Finetuning}

\subsection{Base Model: XTTS v2}
We finetuned XTTS v2 \cite{xtts}, a state-of-the-art multilingual text-to-speech model developed by Coqui. XTTS v2 employs a GPT-style autoregressive architecture that generates discrete audio tokens, which are then decoded into waveforms. The model supports zero-shot voice cloning, allowing it to synthesize speech in a target voice given only a short reference audio clip.
XTTS v2 is pretrained on a large multilingual corpus covering 16 languages, including Arabic. However, the Arabic training data primarily consists of Modern Standard Arabic and Gulf dialects, leaving Egyptian Arabic underrepresented. Our finetuning adapts the model to Egyptian Arabic while preserving its voice cloning capabilities.

\subsection{Finetuning Configuration}
We finetuned the GPT component of XTTS v2 on the NileTTS training set while keeping the DVAE (audio tokenizer) frozen. We largely adopt the default hyperparameters and training setup provided by the Coqui team's finetuning codebase, with minimal modifications. Table \ref{tab:hyperparams} summarizes the key training parameters.

\begin{table}[t]
\centering
\begin{tabular}{lr}
\hline
\textbf{Hyperparameter} & \textbf{Value} \\
\hline
Epochs & 30 \\
Batch Size & 2 \\
Gradient Accumulation Steps & 8 \\
Effective Batch Size & 16 \\
Learning Rate & 5e-6 \\
Optimizer & AdamW \\
Weight Decay & 1e-2 \\
Max Text Length & 400 tokens \\
\hline
\end{tabular}
\caption{Finetuning hyperparameters for XTTS v2 on NileTTS.}
\label{tab:hyperparams}
\end{table}

Our primary modifications to the training pipeline involve integrating Weights \& Biases for experiment tracking and implementing evaluation metrics—including Word Error Rate, Character Error Rate, and Speaker Similarity—computed periodically during training to monitor convergence and enable checkpoint selection.

\section{Experiments and Results}

\subsection{Evaluation Setup}
We evaluate our finetuned NileTTS model against the baseline XTTS v2 model to measure improvements in Egyptian Arabic synthesis quality. The baseline is the pretrained XTTS v2, which includes Arabic but primarily covers Modern Standard Arabic and Gulf dialects.

We use the following evaluation metrics, computed on the held-out evaluation set:
\begin{itemize}
    \item \textbf{Evaluation Loss}: Combined text and mel-spectrogram cross-entropy loss as defined by the XTTS architecture.
    \item \textbf{Word Error Rate (WER)}: We synthesize speech from text, transcribe it using Whisper Large \cite{whisper}, and compute WER against the original text. Lower WER indicates higher intelligibility.
    \item \textbf{Character Error Rate (CER)}: A finer-grained intelligibility metric computed at the character level.
    \item \textbf{Speaker Similarity}: Cosine similarity between ECAPA-TDNN \cite{ecapa} speaker embeddings of synthesized and reference audio. Higher similarity indicates better voice cloning.
\end{itemize}

\subsection{Results}

\begin{figure*}[t]
\centering
\includegraphics[width=\textwidth]{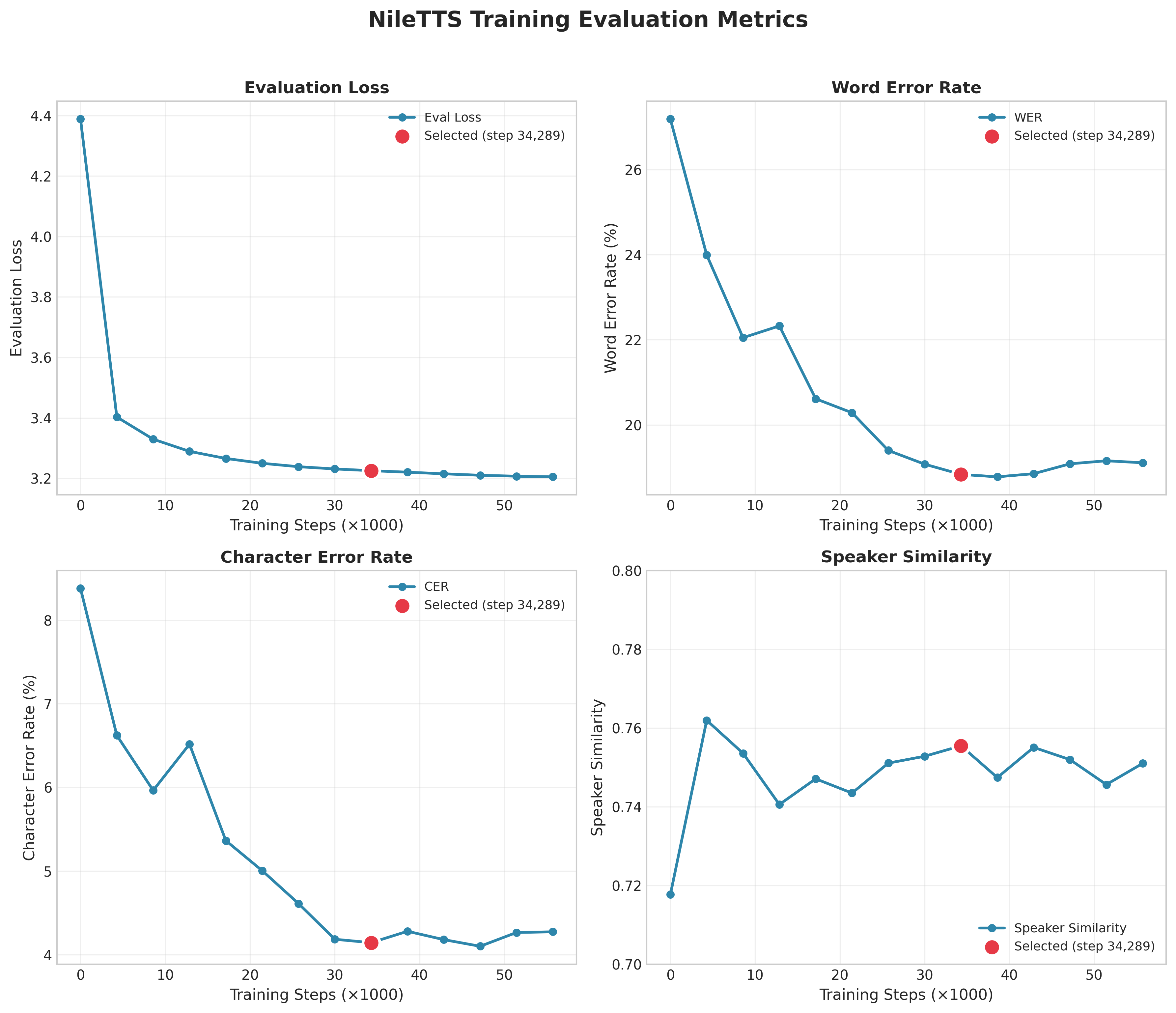}
\caption{Evaluation metrics throughout training: (a) Evaluation Loss, (b) Word Error Rate, (c) Character Error Rate, (d) Speaker Similarity. The red marker indicates the selected checkpoint at step 34,289 (epoch 8).}
\label{fig:training-curves}
\end{figure*}

Figure \ref{fig:training-curves} illustrates the progress of evaluation metrics throughout training. All metrics show rapid improvement in early training, with loss decreasing and intelligibility metrics (WER, CER) improving substantially within the first 20,000 steps. Beyond this point, metrics begin to look more horizontal, indicating diminishing returns from continued training.

\textbf{Checkpoint Selection.} Although we initially planned for 30 epochs of training, we observe that after approximately 8 epochs (around 35,000 steps), the evaluation metrics stabilize with minimal further improvement. Training was stopped after 13 epochs (55,719 steps) due to this reason. We select the checkpoint at step 34,289 (epoch 8), which achieves a strong balance across all metrics. To validate this selection, we synthesized 50 randomly sampled utterances from the evaluation set and conducted manual listening evaluation. The synthesized speech demonstrated natural prosody, accurate pronunciation of Egyptian Arabic phonemes, and consistent preservation of speaker identity, confirming the checkpoint's suitability for release.

Table \ref{tab:results} presents the final comparison between the baseline XTTS v2 model and our finetuned NileTTS model.

\begin{table}[t]
\centering
\begin{tabular}{lccc}
\hline
\textbf{Model} & \textbf{WER} $\downarrow$ & \textbf{CER} $\downarrow$ & \textbf{Spk Sim} $\uparrow$ \\
\hline
XTTS v2 & 26.8\% & 8.1\% & 0.713 \\
NileTTS & \textbf{18.8\%} & \textbf{4.1\%} & \textbf{0.755} \\
\hline
\end{tabular}
\caption{Comparison of baseline XTTS v2 Baseline and finetuned NileTTS on Egyptian Arabic evaluation set.}
\label{tab:results}
\end{table}

NileTTS achieves a \textbf{29.9\% relative reduction in Word Error Rate} (from 26.8\% to 18.8\%) and a \textbf{49.4\% relative reduction in Character Error Rate} (from 8.1\% to 4.1\%), indicating significantly improved intelligibility for Egyptian Arabic synthesis. Speaker similarity improves from 0.713 to 0.755 (\textbf{+5.9\%}), demonstrating better voice cloning.

These results confirm that finetuning on dialect-specific data yields substantial improvements in TTS quality, even when the base model already supports the target language family. We release the NileTTS model weights publicly on Hugging Face to serve as a foundation for future Egyptian Arabic speech synthesis research.

\section{Discussion and Limitations}

\subsection{Dataset Limitations}

While NileTTS represents a significant resource for Egyptian Arabic TTS, limitations should be acknowledged. First, the dataset contains only \textbf{two speakers} (one male, one female), which limits speaker diversity. TTS models trained on limited speaker data may not generalize well to synthesizing voices with different characteristics. Future work should expand the dataset with additional speakers to improve voice diversity.

Second, our dataset is constructed from \textbf{synthetically generated audio} rather than recordings of human speakers. Even though the used audio synthesis tool produces high-quality Egyptian Arabic speech with natural prosody. However, our evaluation results suggest that the synthetic data is sufficient for training effective TTS models, and the pipeline's reproducibility enables future expansion with additional synthetic or natural data.

Third, although we cover three domains (medical, sales, and general conversations), certain \textbf{specialized domains} such as news broadcasting, poetry, or technical content are not represented. Expanding domain coverage would improve the model's versatility.

\subsection{Evaluation Limitations}

Our evaluation relies on \textbf{automatic metrics} (WER, CER, Speaker Similarity) rather than formal human evaluation studies such as Mean Opinion Score (MOS) assessments. While automatic metrics correlate with perceived quality, they do not fully capture subjective aspects like naturalness, expressiveness, or listener preference. We mitigate this limitation through manual listening evaluation of synthesized samples, but a comprehensive human evaluation study remains valuable future work.

Additionally, WER and CER are computed using Whisper Large as the transcription model. While Whisper performs well on Egyptian Arabic, transcription errors from the ASR system may introduce noise into these metrics. Moreover, since Whisper is also used to generate the dataset transcripts, this setup may introduce a form of \textbf{ASR self-consistency bias}, potentially inflating evaluation scores. Future work should evaluate WER and CER using alternative ASR models and include a small human-verified transcription subset to improve robustness.

Similarly, speaker similarity is computed using cosine similarity between ECAPA-TDNN speaker embeddings. As the same embedding architecture is also used in the pipeline, this may introduce a degree of \textbf{model-specific bias} in the speaker similarity scores. Future work will explore the use of additional speaker embedding models for confirmation and incorporate human verification to further validate speaker similarity assessments.

\subsection{Synthetic-to-Real Generalization}

A common concern with synthetic speech datasets is whether models trained on them generalize to real human speech. Although NileTTS is built from synthetically generated audio, the text content is written in authentic Egyptian Arabic, and the synthesis preserves key dialectal properties such as pronunciation, intonation, and prosodic patterns. These aspects are essential for learning dialect-specific speech characteristics.

In low-resource settings, synthetic speech has been shown to be a practical and effective training signal when natural data is limited. In this sense, NileTTS provides a scalable source of Egyptian Arabic speech data that captures core linguistic properties of the dialect and can complement future datasets based on real human recordings.

\subsection{Future Work}

Several directions could extend this work:
\begin{itemize}
    \item \textbf{Speaker expansion}: Adding more speakers with diverse voice characteristics, ages, and speaking styles.
    \item \textbf{Other Arabic dialects}: Applying the synthetic data pipeline to other under-resourced Arabic varieties.
    \item \textbf{Human evaluation}: Conducting formal MOS studies to complement automatic metrics.
    \item \textbf{Robust evaluation}: Evaluating WER, CER, and speaker similarity using multiple independent models and human-verified subsets.
\end{itemize}

\section{Conclusion}

We presented \textbf{NileTTS}, a large-scale Egyptian Arabic text-to-speech dataset and finetuned model. Our dataset comprises 38 hours of transcribed Egyptian Arabic speech from two speakers across medical, sales, and general conversation domains. We introduced a novel \textbf{synthetic data generation pipeline} that leverages large language models for content creation, neural audio synthesis for speech generation, and automatic transcription with speaker diarization—offering a reproducible and scalable approach for creating TTS datasets for low-resource dialects.

By finetuning XTTS v2 on NileTTS, we achieved substantial improvements over the baseline Arabic model: \textbf{29.9\% relative reduction in Word Error Rate}, \textbf{49.4\% reduction in Character Error Rate}, and \textbf{5.9\% improvement in speaker similarity}. These results demonstrate that dialect-specific finetuning significantly enhances TTS quality for underrepresented language varieties.

We publicly release the NileTTS dataset, model weights, and pipeline code to facilitate further research in Egyptian Arabic speech synthesis. We hope this work contributes to closing the resource gap for Arabic dialects and inspires similar efforts for other low-resource language varieties.

\bibliography{custom}
\end{document}